\documentclass{article}
\usepackage{spconf,amsmath,graphicx}
\usepackage{amssymb}
\usepackage{booktabs,multirow}
\usepackage{enumitem}
\setlist{nosep,leftmargin=14pt}
\nocite{IEEEexample:BSTcontrol}

\title{BiSe-UNet: A Lightweight Dual-Path U-Net with Attention-Refined Context for Real-Time Medical Image Segmentation}

\name{M. Iffat Hossain$^{1}$ \qquad Laura J. Brattain$^{2}$ \thanks{This work has been submitted to the IEEE for possible publication. Copyright may be transferred without notice, after which this version may no longer be accessible.}}

\address{
$^{1}$ Electrical and Computer Engineering, University of Central Florida \\
$^{2}$ Department of Medicine, University of Central Florida College of Medicine
}

\begin{document}
\maketitle

\begin{abstract}
During image-guided procedures, we often need real-time image segmentation. This requires lightweight AI models that can run on resource-constrained devices. One use case is endoscopy-guided colonoscopy, during which polyps are detected at real-time. The Kvasir-Seg dataset, a publicly available benchmark for this task, features 1,000 high-resolution endoscopic images of polyps with corresponding precise pixel-level segmentation masks. Achieving the required real-time inference speed for clinical use in resource-constrained environments necessitates the deployment of highly efficient and lightweight network architectures. However, most models remain too computationally intensive for embedded deployment. The existing lightweight architectures, though faster, often have reduced spatial precision and contextual understanding—resulting in degraded boundary quality and reduced diagnostic reliability. To overcome these challenges, we present BiSe-UNet, a lightweight dual-path U-Net that integrates an attention-refined Context Path with a shallow Spatial Path fused for detailed feature preservation, followed by a depthwise-separable decoder for efficient reconstruction. Evaluated on the Kvasir-SEG dataset, BiSe-UNet achieves competitive Dice and IoU scores while sustaining real-time throughput exceeding 30 FPS on Raspberry Pi 5, demonstrating its effectiveness for accurate, lightweight, and deployable medical image segmentation on edge hardware.
\end{abstract}


\section{Introduction}

The integration of computer-aided diagnosis (CAD) is crucial for image-guided robotic procedures in interventional medicine, where real-time semantic segmentation enables precise localization and guidance. This pixel-level analysis is essential for tasks such as colorectal polyp delineation. While high accuracy in segmentation is essential for patient safety, clinical deployment demands real-time performance ($\geq$30~FPS) on resource-constrained hardware.

Over the past decade, U-Net~\cite{Ronneberger2015} and its variants have enabled high accuracy segmentation through residual~\cite{ResUNet}, dense~\cite{UNetPP}, and attention mechanisms~\cite{AttUNet}. Despite the advances, Convolutional Neural Network (CNN)-based models remain over-parameterized and computationally heavy. While transformer-based architectures~\cite{Dosovitskiy2020, Segformer} improve global context modeling, they incur quadratic attention costs, limiting real-time deployment. Hybrid CNN–Transformer models~\cite{TransUNet, UTNet} integrate local and global features yet add training and inference overhead~\cite{QuantizationSurvey}. Efficiency-focused designs like Fast-SCNN~\cite{FastSCNN} and BiSeNet~\cite{Yu2018} prioritize speed via dual-branch structures; however, balancing inference speed and structural fidelity remains challenging~\cite{Chen2018}. Models like HarDNet~\cite{HarDNet2019} speed up processing by flattening features into fewer channels, but result in loss of fine boundary details and lead to lower segmentation accuracy. Moreover, synchronization and memory overheads hinder real-time performance on embedded devices such as Jetson~\cite{jetson} and Raspberry Pi~\cite{Sze2017}.

\begin{figure*}[t]
	\centering
    \includegraphics[width=\textwidth]{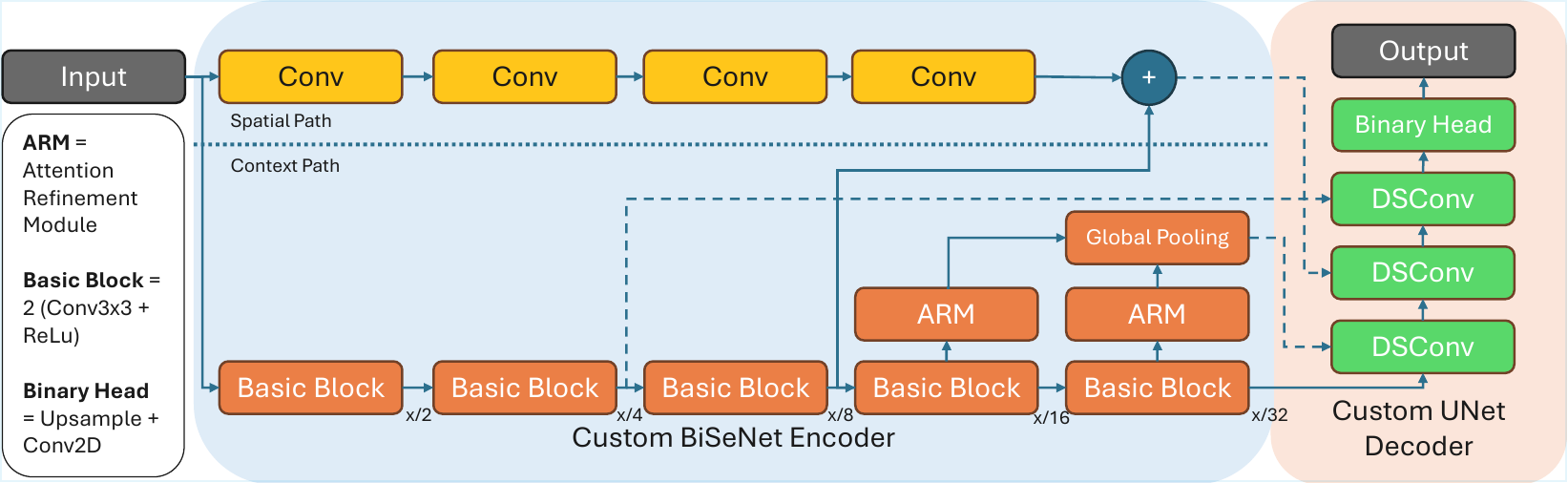}
	\caption{The workflow shows BiSe-UNet extracting spatial and contextual features in two paths, merging them once for efficient integration, and decoding with a lightweight DSConv module to retain boundaries with minimal computation.}
	\label{fig:workflow}
\end{figure*}

To address these limitations, we propose \textbf{BiSe-UNet}, a U-Net variant inspired by the spatial–context separation of BiSeNet but optimized for real-time medical segmentation on constrained hardware. Our model pairs an attention-enhanced encoder with a lightweight decoder and uses combined context–spatial features as skip connections, which gives the decoder direct access to early image details so that boundaries remain clear. The overall framework is shown in Figure~\ref{fig:workflow}, where features from both paths are merged before being decoded through a DSConv-based reconstruction module.

Our core contributions, addressing the need for deployable, real-time medical segmentation, are as follows:

\begin{enumerate} 
\item We introduce \textbf{BiSe-UNet}, a novel dual-path lightweight U-Net designed for real-time medical segmentation, integrating an Attention-Refined Context Path with a shallow Spatial Path for efficient and accurate feature extraction.
\item We utilize a Depthwise-Separable Convolution (DSConv) decoder to minimize computational load (MACs \& parameters) while preserving high segmentation quality \cite{Howard2017, Sandler2018}.
\item BiSe-UNet achieves a superior Pareto trade-off between accuracy (Dice/IoU) and speed, enabling reliable real-time inference (30+ FPS) on embedded hardware, specifically demonstrating deployment on the Raspberry Pi 5. 
\end{enumerate}

\section{Materials and Methods}
\label{sec:method}

\subsection{Dataset Description \& Preparation}
We use \emph{Kvasir-SEG} dataset, which is a dataset of gastrointestinal polyp images and corresponding segmentation masks \cite{Jha2020}. It contains 1,000 high-resolution colonoscopy images with pixel-level polyp annotations collected from 47 patients. Each frame represents real endoscopic scenes with diverse polyp shapes, textures, and illumination conditions, making it suitable for evaluating medical segmentation performance. Images and masks are paired by filename stem; non-matching files are skipped. The dataset is split into train/val/test with proportions $70\%/15\%/15\%$ using a fixed seed (\verb|SEED=42|). Images are normalized using ImageNet statistics (\textmu = [0.485, 0.456, 0.406], $\sigma$=[0.229, 0.224, 0.225]) and converted to tensors with all channels; masks are returned as $1{\times}H{\times}W$ float tensors. Training-time augmentation includes horizontal/vertical flips, 90°/180°/270° rotations, and brightness/contrast jitter, each controlled by configuration probabilities.

\subsection{Methods}
\label{sec:methods}
Figure~\ref{fig:workflow} shows the architecture of \textbf{BiSe-UNet}. It consists of three sections: Context Path (CP), Spatial Path (SP), and Lightweight UNet Decoder. The Context Path (CP) downsamples the input feature maps to progressively coarser spatial resolutions. In parallel, a shallow Spatial Path (SP) preserves fine-grained structural detail. Finally, a lightweight U-Net decoder using bilinear upsampling and depthwise separable (DSConv) fusions reconstructs the segmentation mask at full resolution.

To ensure fair and comprehensive evaluation, both accuracy and efficiency metrics were considered. The Dice coefficient and Intersection over Union (IoU) assess segmentation precision and boundary overlap between predictions and ground truth masks. Inference speed, measured in frames per second (FPS), indicates the model’s real-time capability across hardware platforms, while memory shows the resource consumption while completing the inference. Finally, the Multiply–Accumulate Operations (MACs) and total trainable parameters (Params) reflect the computational and memory efficiency of each architecture.

Given input $x\!\in\!\mathbb{R}^{B\times C\times H\times W}$, CP yields multi-scale features, generating $x_{/4}$, $x_{/8}$, $x_{/16}$, $x_{/32}$ where each fraction denotes the reduction ratio in height and width after successive stride-2 convolutions. Following BiSeNet, we employ \emph{Attention Refinement Modules (ARM)} at $x_{/16}$ and $x_{/32}$, each consisting of a local $3{\times}3$ conv+Batch Normalization (BN)+ReLU and a squeeze-style gate (global average pooling, $1{\times}1$ conv+BN, sigmoid). A lightweight global-context head on ARM of $x_{/32}$ is upsampled, and merged with ARM of $x_{/16}$ forming $x^{\mathrm{ref}}_{/16}$. Thus, The CP exposes skip connections at $x_{/4}$, $x_{/8}$, and $x^{\mathrm{ref}}_{/16}$ resolutions for multi-scale fusion.

The SP is a shallow, high-resolution stream ending at $s_{/8}$ from input $x$: a $7{\times}7$ stride-2 layer followed by two $3{\times}3$ stride-2 layers and a $1{\times}1$ projection. The $1{\times}1$ projection is a single-channel convolution that aligns feature dimensions without altering spatial size. Its output $s_{/8}\!\in\!\mathbb{R}^{B\times C_b\times H/8\times W/8}$ captures edges and fine detail. We concatenate $s_{/8}$ with the CP $x_{/8}$ and project back for skip connection instead of $x_{/8}$ using a $1{\times}1$ conv+BN+ReLU:
\[
x'_{/8} = \phi\big([\,x_{/8}\,\|\,s_{/8}\,]\big), \quad \phi:\;1{\times}1\text{ conv+BN+ReLU}.
\]

The decoder applies a sequence of upsampling and fusion steps, using the context-path skip tensors and DSConv blocks. Each DSConv block is constructed from a depthwise $3{\times}3$ convolution and a pointwise $1{\times}1$ convolution, both followed by BN and an in-place ReLU activation. The decoder first upsamples $x_{/32}$ and merges it with $x^{\mathrm{ref}}_{/16}$ through a Depthwise Separable Convolution (DSConv) block. The resulting features are then upsampled again and fused with the mid-level skip tensor $x'_{/8}$ using a second DSConv block. A further upsampling step enables fusion with the higher-resolution skip $x_{/4}$ via a third DSConv block. After one last upsampling, a $1{\times}1$ prediction head generates the logits, which are finally interpolated back to full image size. 

We use standard segmentation losses (e.g., Dice, BCE, or a weighted sum), standard data augmentation, and input sizes aligned with deployment (e.g., $256^2$ or $320^2$). Optimizer and LR schedule follow common practice and converted to ONNX models according to the target hardware constraints.

\section{Experimental Results}
\label{sec:results}

\begin{figure}[t]
  \centering
  \includegraphics[width=\columnwidth]{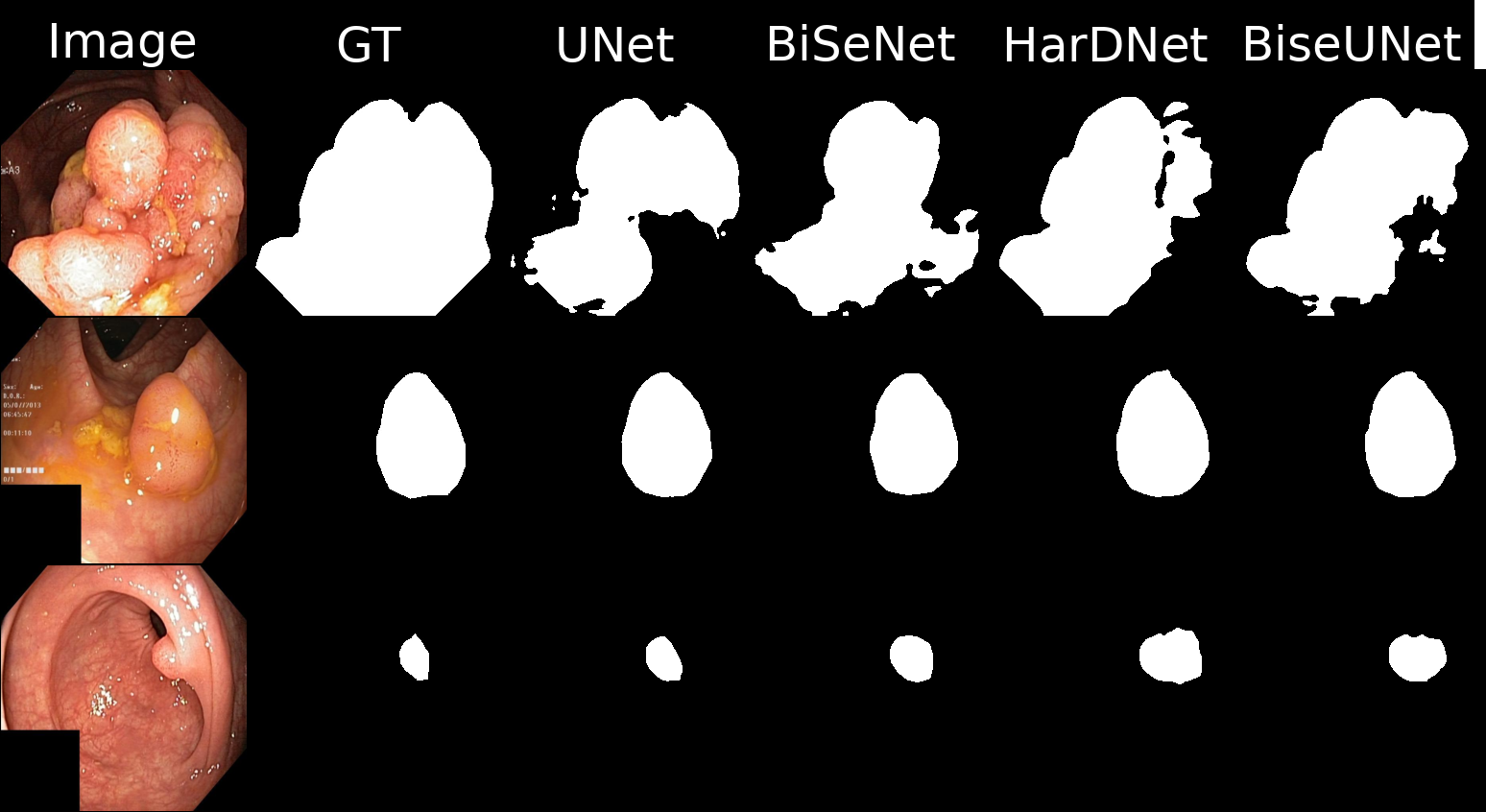}
  \caption{Polyp Segmentation Performance Comparison -  BiSe-UNet against popular existing segmentation models}
  \label{fig:comparison}
\end{figure}

\begin{table*}[t]
\centering
\caption{Comparison of segmentation models on CUDA (GTX\,1080\,Ti) and Raspberry\,Pi\,5.
Shared metrics include model size, computational cost, and segmentation accuracy. 
FPS and Memory indicate device-specific throughput and runtime footprint.}
\label{tab:main}
\resizebox{\textwidth}{!}{%
\begin{tabular}{lcccccccccc}
\toprule
\textbf{Model} & \textbf{Params (M)} $\downarrow$ & \textbf{MACs (G)} $\downarrow$ & 
\textbf{Dice} $\uparrow$ & \textbf{IoU} $\uparrow$ &
\multicolumn{2}{c}{\textbf{CUDA (GTX\,1080\,Ti)}} & & 
\multicolumn{2}{c}{\textbf{Raspberry\,Pi\,5}} \\
\cmidrule(lr){6-7} \cmidrule(lr){9-10}
 &  &  &  &  & \textbf{FPS} $\uparrow$ & \textbf{Memory (MB)} $\downarrow$ & & \textbf{FPS} $\uparrow$ & \textbf{Memory (MB)} $\downarrow$ \\
\midrule
U-Net (baseline)     & 7.813 & 11.67 & 0.7900 & 0.7000 & 217.91 & 420 & & 2.65  & 300 \\
BiSeNet   & 2.533 & 1.07  & 0.7501 & 0.6595 & 397.37 & 210 & & 30.06 & 160 \\
HarDNet              & 3.809 & 4.46  & 0.7775 & 0.6959 & 232.62 & 360 & & 7.17  & 200 \\
\textbf{BiSe-UNet (Ours)} & \textbf{2.509} & \textbf{0.97} & \textbf{0.7809} & \textbf{0.6961} & \textbf{358.34} & \textbf{240} & & \textbf{30.48} & \textbf{170} \\

\bottomrule
\end{tabular}%
}
\end{table*}

\begin{table*}[t]
\centering
\caption{Ablation on BiSe-Unet variants (Pi5 CPU, batch=1, 256$\times$256).}
\label{tab:BiSe-Unet_ablation_pi5}
\begin{tabular}{l r r r r r r}
\toprule
Model & Params (M) & MACs (G) & FPS (Pi5) & Latency (ms) & Dice & IoU \\
\midrule
CP+SP(/8) +  (Current) & 2.53 & 1.04 & 32.40 & 30.86 & 0.8037 & 0.7174 \\
DSConv encoder & 2.51 & 1.01 & 36.83 & 27.15 & 0.7931 & 0.7052 \\
DSConv Encoder + Decoder & 2.60 & 1.15 & 29.32 & 34.11 & 0.7929 & 0.7051 \\
BiSe-Unet (baseline) & 2.63 & 1.20 & 27.05 & 36.97 & 0.7855 & 0.6974 \\
\bottomrule
\end{tabular}
\end{table*}

\subsection{Result Analysis}
\label{sec:results_analysis}

Figure \ref{fig:comparison} shows alpha mask comparisons with the ground truth, where BiSe-UNet yields the most accurate and well-defined boundaries. Table~\ref{tab:main} presents the quantitative comparison of segmentation performance and efficiency across all models. The proposed BiSe-UNet shows a balanced accuracy and efficiency, achieving a Dice score of 0.7809 with only 2.5 M parameters—comparable to U-Net’s accuracy (0.7900) while reducing computational cost by \textbf{over 90\%} (0.97 G vs. 11.67 G MACs).

Compared with BiSeNet, BiSe-UNet improves Dice by \textbf{+4.1\%} and IoU by \textbf{+5.5\%} using similar parameter counts, highlighting its superior spatial–context fusion. Using CUDA, it maintains 358 FPS, which is \textbf{65\%} faster than U-Net while consuming \textbf{43\%} less memory. On the Raspberry Pi 5, BiSe-UNet sustains 30.5 FPS—nearly \textbf{10×} faster than U-Net and \textbf{4×} faster than HarDNet~\cite{HarDNet2019} with a 40\% lower memory footprint. Overall, the model demonstrates the best balance between accuracy, speed, and resource efficiency among all tested architectures.

\subsection{Ablation Study}
\label{sec:ablation}

\begin{figure}[t]
  \centering
  \includegraphics[width=\columnwidth]{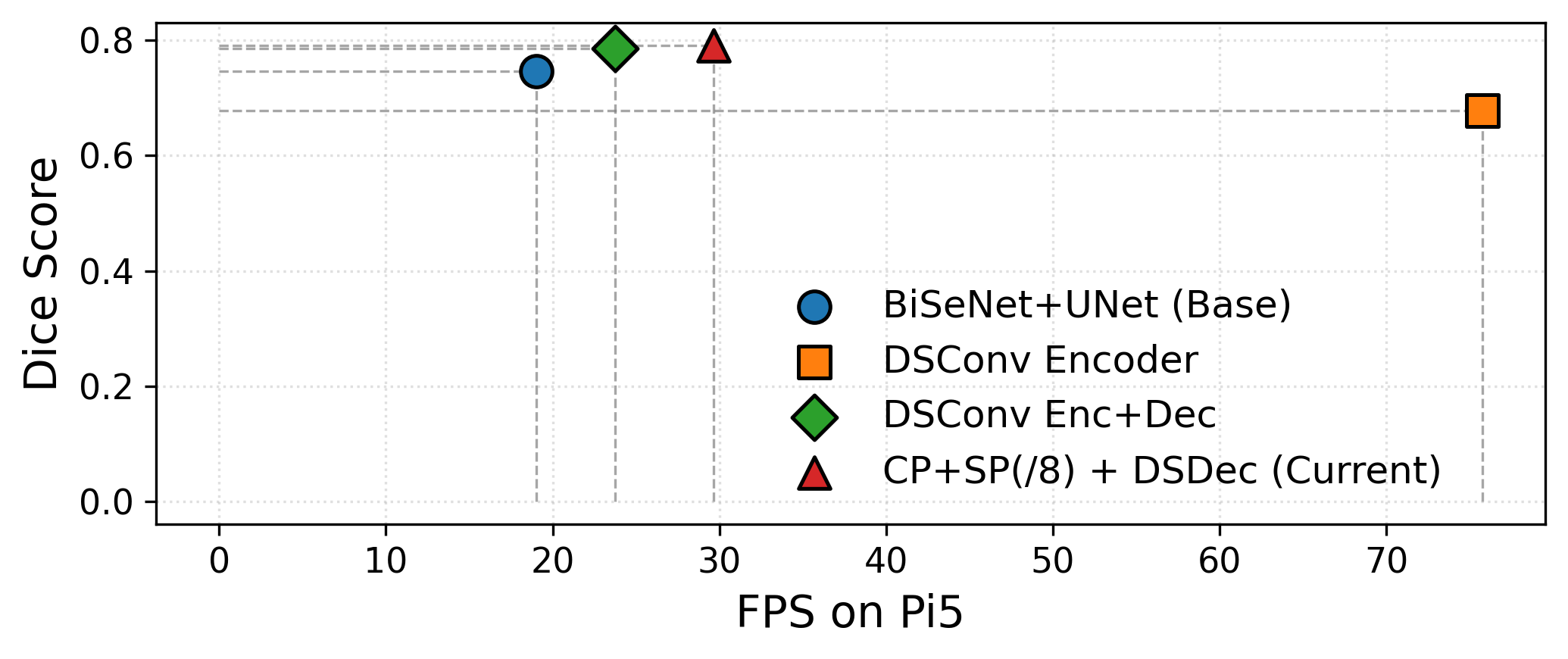}
  \caption{Ablation of BiSe-UNet tested on Raspberry\,Pi\,5. The square (DSConv encoder) improves speed but lowers Dice Score compared to the base model (circle); the diamond (DSConv in both encoder and decoder) maintains high accuracy, while the triangle (final BiSe-UNet with custom skip connections) results in good speed and Dice score.}
  \label{fig:ablation-figure}
\end{figure}

Figure~\ref{fig:ablation-figure} illustrates the progressive improvements from different BiSe-UNet configurations on Raspberry~Pi~5. Table \ref{tab:BiSe-Unet_ablation_pi5} further summarizes the ablation experiments conducted to evaluate each architectural component of \textbf{BiSe-UNet}. Replacing the standard convolutional blocks with depthwise-separable convolutions reduced computation by about 15–20\% with only a minor Dice drop ($\sim 0.79$), confirming the efficiency of lightweight operators. Introducing the dual-path fusion design further improved Dice by +1.3 points and IoU by +1.0 point, highlighting the benefit of parallel spatial–context encoding. The final fusion at the \textit{/8} resolution achieved the highest Dice (0.8037) and IoU (0.7174) with 32 FPS on Raspberry Pi 5—demonstrating an optimal trade-off between accuracy and latency for real-time embedded inference. Overall, the results verify that combining dual-path encoding with early spatial–context fusion and DSConv decoding yields the most balanced configuration of BiSe-UNet.

\section{Discussion}

\label{sec:discussion}

The proposed \textbf{BiSe-UNet} effectively balances accuracy and efficiency by integrating attention-refined contextual encoding with a lightweight dual-path design. The single fusion at the $1/8$ of input $x$ scale preserves fine boundary details without introducing redundant computation, while the depthwise-separable decoder maintains strong segmentation quality at a fraction of the cost of standard U-Net decoders. Experimental results confirm that this selective fusion enables consistent Dice and IoU performance across hardware, achieving real-time inference on edge devices such as Raspberry~Pi~5.

BiSe-UNet demonstrates that careful architectural design can outperform heavier networks. Its stability across resolutions makes it a promising candidate for real-time endoscopic segmentation. Future extensions could explore multi-class segmentation, adaptive quantization, and dynamic input scaling to enhance generalization and further optimize deployment on constrained medical hardware.

\section{Conclusion}

\label{sec:conclusion}

We introduced \textbf{BiSe-UNet}, a compact dual-path U-Net variant that fuses spatial and contextual features using attention refinement and depthwise-separable decoding. The model achieves real-time segmentation with competitive accuracy on datasets while maintaining a minimal parameter footprint. These results highlight that lightweight, hardware-aware architectures can deliver high performance, paving the way for efficient medical AI systems on edge platforms.

\bibliographystyle{IEEEtran}
\bibliography{refs}

\end{document}